\documentclass{article}
\usepackage{spconf,amsmath,graphicx}
\usepackage{paralist,cite}
\usepackage{amssymb,color,amsmath}

\newcommand{\nix}[1]{}

\title{Efficient Web-based Facial Recognition System Employing 2DHOG}
\name{Moataz M. Abdelwahab$^{1}$~~~~~~~~Salah A. Aly$^{2}$~~~~~~~~~Islam Yousry$^{1}$\thanks{Thanks to HajjCoRE, Center of Research Excellence in Hajj~ and~ Umrah at~ Umm Al-Qura University in KSA, agency for funding this work.}}
\address{$^{1}$School of Communications and Information Tech., Nile University, Smart Village, Egypt \\ \{mabdelwahab,islam.yousry\}@nileuniversity.edu.eg\\ $^{2}$Department of Computer Science, Umm Al-Qura University, Makkah, KSA\\  salahaly@uqu.edu.sa  }

\begin{document}

\maketitle

\begin{abstract}
In this paper, a system for facial recognition to identify missing and found people in Hajj and Umrah is described as a web portal. Explicitly, we present a novel algorithm for recognition and classifications of facial images based on applying 2DPCA to a 2D representation of the Histogram of oriented gradients (2D-HOG) which maintains the spatial relation between pixels of the input images. This algorithm allows a compact representation of the images which reduces the computational complexity and the storage requirments, while maintaining the highest reported recognition accuracy. This promotes this method for usage with very large datasets. Large dataset was collected for people in Hajj. Experimental results employing ORL, UMIST, JAFFE, and HAJJ datasets confirm these excellent properties.
\end{abstract}
\begin{keywords}
Facial recognition, 2DPCA, 2D-HOG
\end{keywords}

\section{Introduction}\label{sec:intro}

Human facial recognition seems at the first glance to be a challenging task due to several reasons such as changes in haircuts, beards, wearing glasses etc. Also human faces depend on people's race and culture. On the other hand, more than ten million people visits KSA yearly, among them several thousand got missing due to the huge crowds in the two holy cities of Makkah and Madinah.

\begin{figure}[t]
  \begin{center}
  \includegraphics[width=8.8cm,height=5.5cm]{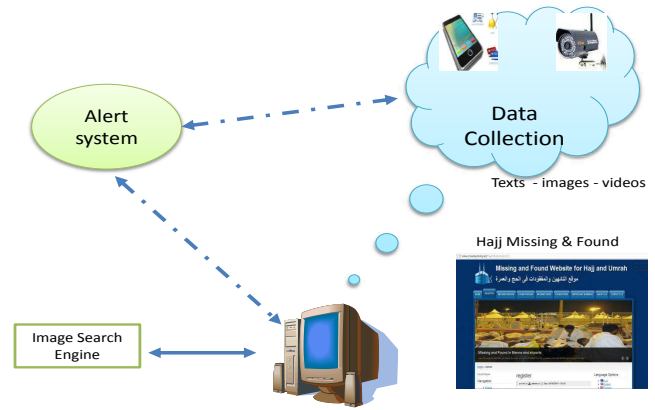}
  \caption{Crowd-Sensing system. The system consists of data collectors (mobile, cameras, PCs), main server, search engine, and an alerting system.}\label{fig:hajjmfsys}
  \end{center}
\end{figure}

We propose a Crowd-Sensing system as shown in Fig.~\ref{fig:hajjmfsys} that is composed of a web portal including a facial recognition algorithm. The Crowd-Sensing system is established to support the existing efforts to manage the crowds and solve the missing and found problem during Hajj and Umrah seasons in KSA. The goal of this Crowd-Sensing system is to use techniques from computer vision and image processing to develop a portal website for Hajj missing and found people~\cite{crowdsensing},~\cite{hajjdataset}.

Many algorithm based on principal component analysis (PCA) have been applied to the facial recognition problem. The main idea of PCA is to find the vectors that best account for the distribution of facial images within the entire image space. In 1991 Turk and Pentland~\cite{Turk91v2},~\cite{Turk91}  developed the Eigenfaces method based on the principal component analysis (PCA). Recently Yang et. al.~\cite{yang2004} proposed the two dimensional PCA (2DPCA) technique, which has many advantages over the PCA method. It is simpler for image feature extraction, better in recognition rate and more efficient in computation complexity. In this contribution we are presenting an efficient method based on applying 2DPCA to a proposed 2D representation of the HOG (2D-HOG) where it maintains the spatial relation between pixels. In addition, this algorithm allows compact representation of the images which yields excellent recognition speed and storage requirements, while maintaining the highest reported recognition accuracy.

\begin{figure}[t]
  \begin{center}
  \includegraphics[width=8.5cm,height=5cm]{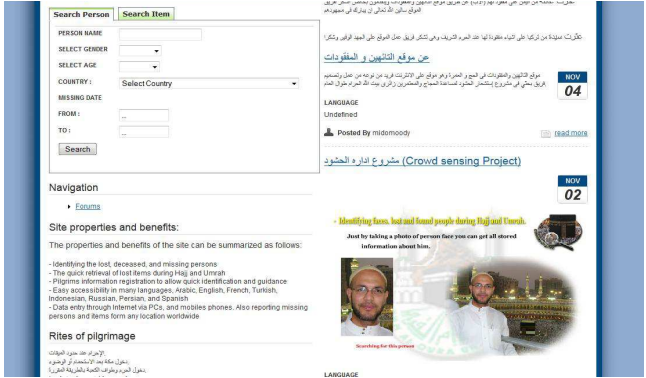}
  \caption{HajjMF portal interface website developed by Crowd-Sensing.net team}\label{fig:hajjmf2}
  \end{center}
\end{figure}

\section{Crowd-Sensing System Description}\label{sec:systemdescription}
In this section, we describe the Crowd-Sensing system for recognizing missing and found people during Hajj and Umrah seasons in KSA. The main idea is that pilgrims send images of their missing/found individuals and the developed web portal as shown in Fig.~\ref{fig:hajjmf2}, can detect and recognize their faces. Hence then, all person information can be known easily. The Crowd-Sensing tool can be used for road management, media, surveillance, crowd estimation, crowd flow monitoring, crowd data collection (speed, size, direction, and stampede) and flow jams reduction.

The proposed system has several advantages including a tool for identifying missing and found people during dense crowded situations. It also contains a search engine for all missing objects. Furthermore, it contains a database, which can save up to three millions records. The system works as follows:
\begin{compactenum}[i)]
\item
A database will be setup for all people observing Hajj and Umrah.
\item
A person photo will be uploaded to the website. A computer program will be used to  recognize the missing/found person employing the novel facial recognition algorithm presented in this paper.
\item The system will be capable of recognizing human facial images in the presence of various anomalies.
\item It worth to mention that, In addition to facial recognition application presented in this paper the system will be able to detect, count recognize crowds and people behavior. Our method will be robust, and not subject to failure against pollution, temperatures, or any other atmosphere changes.
\end{compactenum}

\begin{figure}[t]
\begin{center}
   \includegraphics[width=8.2cm,height=6.5cm]{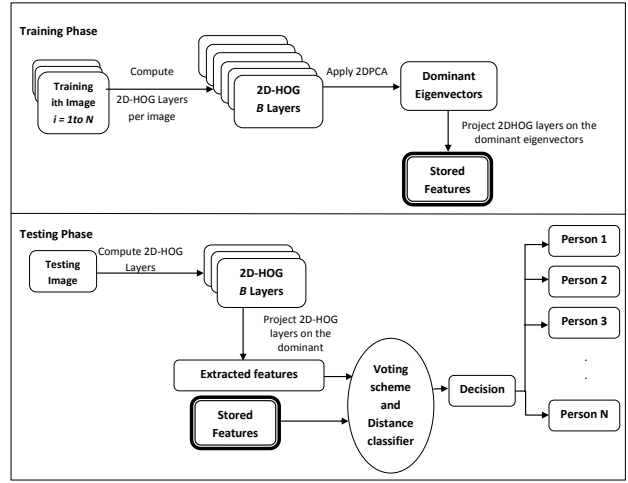}
  \caption{Face Recognition System Description. Figures (a) and (b) show steps of the facial training and testing detection algorithms.}\label{fig:traintest}
  \end{center}
\end{figure}


\section{PROPOSED ALGORITHM (2DHOG /2DPCA)}\label{sec:proposedalg}
In 2006 Dalal and Triggs~\cite{dalal05} introduced a single window human detection algorithm with excellent detection results. This method uses a dense grid of Histograms of Oriented Gradients (HOG) for feature extraction and using linear Support Vector Machine (SVM) for classification. The HOG representation has several advantages. It captures edge or gradient structure that is the main feature of local shape, and it is robust for illumination changes, invariant for human clothes, and background changes.

 In this paper, a novel algorithm for human face recognition is introduced, where 2DPCA is applied to Histogram of Oriented Gradients (HOG) represented into $n$ orientation bins that are layers in $2D$ format. Feature extraction process can be either in spatial or transform domains ~\cite{Abdelwahab07},~\cite{Abdelwahab06}.

\begin{figure}[t]
  \begin{center}
  \includegraphics[width=8cm,height=6cm]{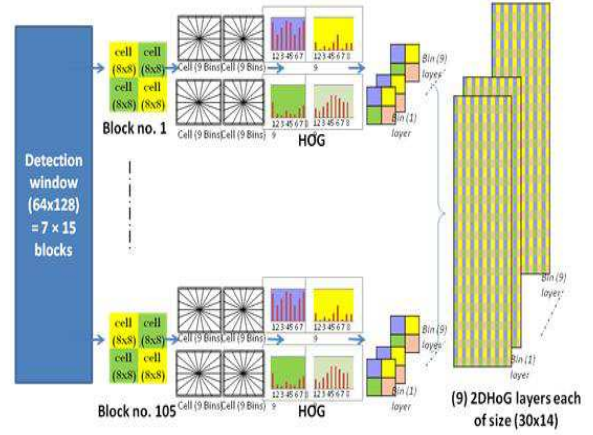}
  \caption{An example for the proposed 2DHoG Algorithm}\label{fig:hog}
  \end{center}
\end{figure}

 The technical contributions of this part are two folds, first representing the HOG features in 2D format so that the relation between HOG features is maximized. Second, each bin will contribute to the decision separately in a parallel structure which reduces the computational time while achieving excellent accuracy levels, as other state-of-the-art approaches. The 2D-HoG B-bin features extraction, consists of a parallel structure (layers) as shown in Fig.~\ref{fig:hog}.

 The $B$ bins, where each bin represents one of the desired angles, are arranged in $B$ matrices, where spatial relations are maintained. Each bin will be dealt with separately in a parallel structure. The stored feature vector representing the training image will be formed by the concatenation of features extracted from all $B$ bins. In the testing mode, a minimum distance classifier combined with a voting scheme will be used to measure the similarity between the features representing testing and training images.

\begin{figure}[t]
  \begin{center}
  \includegraphics[width=8.7cm,height=6cm]{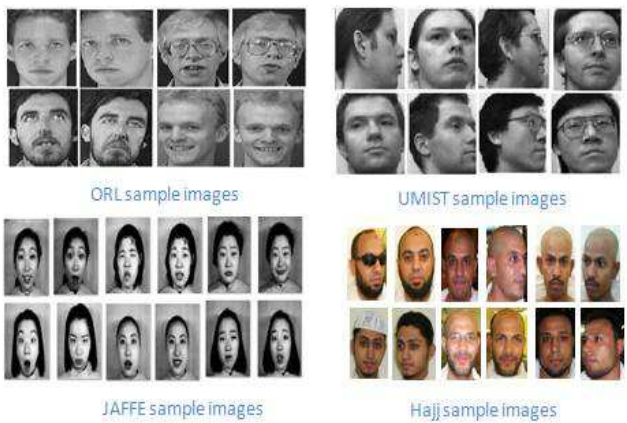}
  \caption{Samples  of facial images (a) ORL, (b) UMIST, (c) JAFFE, (d) HAJJ}\label{fig:DPCA-model2}
  \end{center}
\end{figure}

\noindent\textbf{ORL Dataset:} The ORL dataset consists of 400 images of 40 different individuals (10 images for each persons), where pose and facial expressions are varying, Fig.~\ref{fig:DPCA-model2}a.

\noindent\textbf{UMIST Dataset:} The UMIST dataset consists of 564 images of 20 individuals (mixed race/gender/appearance). Each individual is shown in a range of poses from profile to frontal views Fig.~\ref{fig:DPCA-model2}b.

\begin{figure}[!h]
  \begin{center}
  \includegraphics[width=8cm,height=12.5cm]{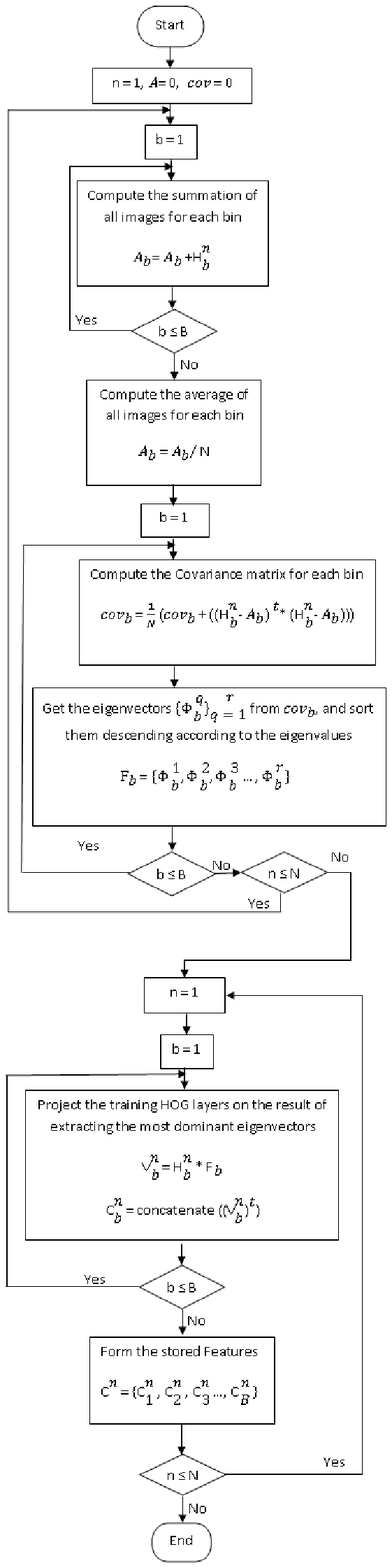}
  \caption{Flowchart for the training mode}\label{fig:trainingAlgo2}
  \end{center}
  \begin{center}
  \includegraphics[width=7.5cm,height=7cm]{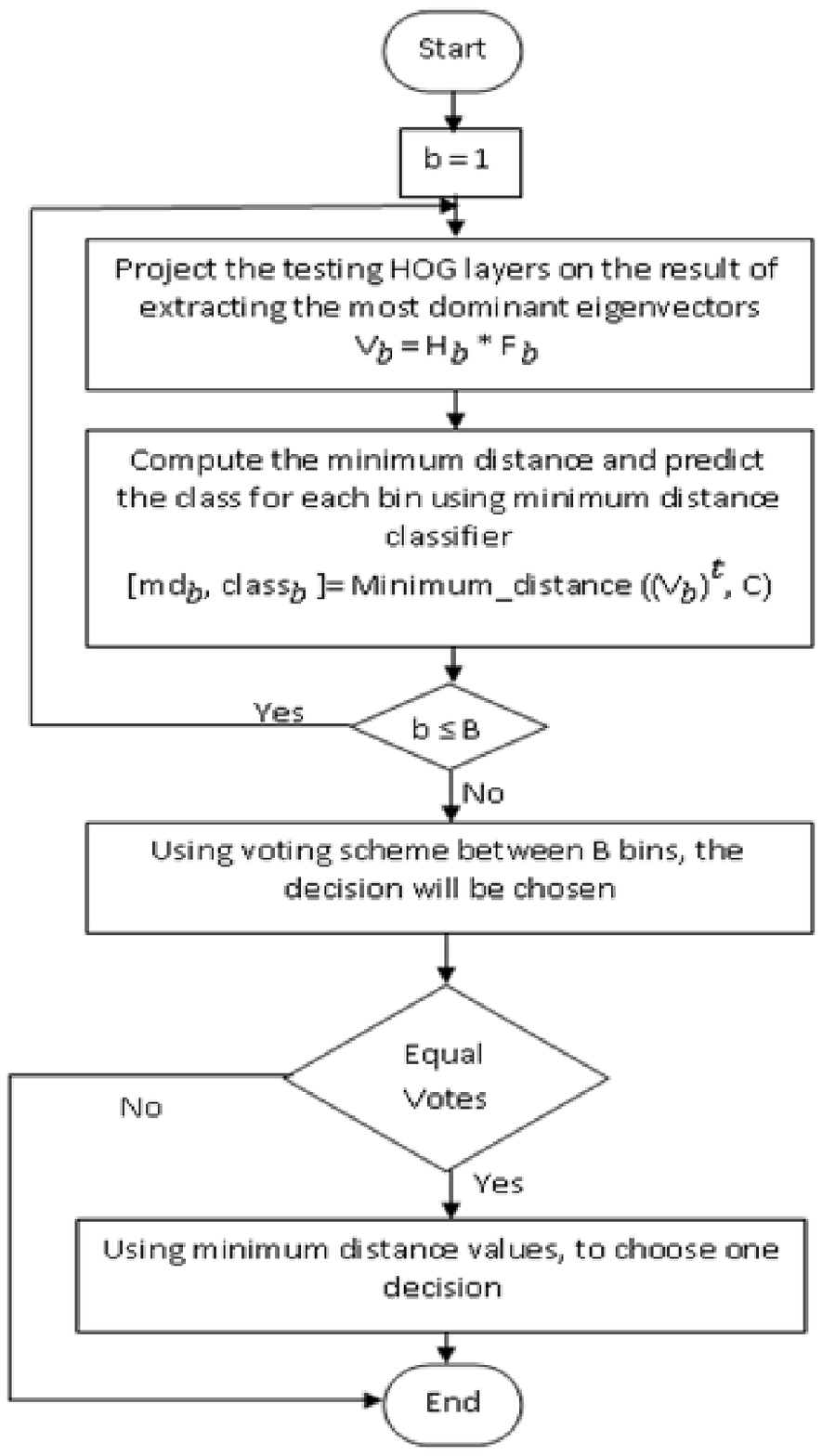}
  \caption{Flowchart for the testing mode}\label{fig:testingAlgo}
  \end{center}
\end{figure}

\noindent\textbf{JAFFE Dataset:} The JAFFE database consists of 10 persons, The images per person vary from 20 to 23  , where facial expressions are varying, Fig.~\ref{fig:DPCA-model2}c.

\noindent\textbf{Hajj Dataset:} HAJJ dataset was taken randomly during Hajj and Umrah seasons. It consists of $100$ different individuals (images per person vary from $3$ to $7$), where pose, facial expressions, and appearance (shaved hair, wearing glasses, etc.) are varying. Samples for these datasets are shown in Fig.~\ref{fig:DPCA-model2}d.

\section{EXPERIMENTAL RESULTS AND ANALYSIS}\label{sec:results}

Three experiments were conducted on the ORL, UMIST, and JAFFE dataset. In the first experiment, five images were used for training and the remaining images were used for testing. In the second experiment three images were used for training and the remaining images were used for testing. In the third experiment the leave-one-out strategy was used. Experimental results were compared with those obtained by PCA and 2DPCA methods.

In our experiments, the recognition system works as follows. The image of size $112x96$ in the spatial domain or of size $56x48$ in the Discrete Wavelet Transform (DWT) domain was used. The cell size is $4x4$ pixels, and the block size is $2x2$ cells, with no overlapping. The ten eigenvectors corresponding to the ten largest eigenvalues of the covariance matrix were obtained. Consistent accuracies were obtained in both the spatial and transform domain. The storage and computational requirements were reduced in the transform (DWT) domain. Results shown in the following tables were obtained employing DWT.

As shown in Table~\ref{table:result1}, the recognition accuracy has improved compared to the excellent recognition accuracy of the 2DPCA method.  From Table~\ref{table:result2}, the computational complexity is less than those required by PCA, and 2DPCA methods. Table~\ref{table:result3},  shows  comparable storage requirements to those required by 2DPCA. method.

In addition, two experiments were conducted on the Hajj dataset with the same parameters used in the previous experiments. In the first experiment, two images were used for training and one image was used for testing for each individual. It worth to note that these images are with different orientations and appearance. In the second experiment, one frontal image was used for training, another frontal image with different appearance and scale was used for testing.  In the first experiment, an accuracy of 92.5\% was obtained, while in the second experiment, 100\% accuracy was achieved. These excellent properties promote our proposed method for usage with large scale datasets for real time applications.

\begin{table}[h]
\centering
\caption{Recognition accuracy for experiment I employing  PCA, 2DPCA, and 2D-HOG/2DPCA methods.
}	\label{table:result1}
\begin{tabular}{|c|c|c|c|}
 \hline
  &	PCA	&2DPCA&	    2D-HOG/2DPCA\\
  \hline
ORL	&92.5\%	&96\%	&97\% \\
UMIST	&85\%	&88\%&	90.35\% \\
JAPANESE	&93.5\%	&96.00\%	&100.00\% \\
  \hline
\end{tabular}
\end{table}

\begin{table}[h]
\centering
\caption{
Computational Complexity (CPU 2.3 GHz, RAM 1G) for experiment I on ORL dataset employing PCA, 2DPCA, and 2DHOG/2DPCA.
	}\label{table:result2}
\begin{tabular}{|c|c|c|c|}
  \hline
 & PCA&	2DPCA&	2DHOG/2DPCA\\
 \hline
Training Mode	&101sec&	14sec&	10.4 sec\\
Testing Mode	&2.1 sec&	1.5 sec	&0.82 sec\\
  \hline
\end{tabular}
\end{table}
\begin{table}[h]
\centering
\caption{
 Storage requirements for experiment I employing 2DPCA, and 2DHOG/2DPCA.
	}\label{table:result3}
\begin{tabular}{|c|c|c|c|}
  \hline
 & ORL	&UMIST&	JAFFE\\
 \hline
2DPCA	&112x12&	112x12&	112x12\\
2DHOG/2DPCA	&14x12x8&	14x12x20&	14x12x11\\
 \hline
\end{tabular}
\end{table}

\begin{table}[h]
\centering
\caption{
Comparison between Recognition accuracy for experiments I, II, and III employing 2DHOG/2DPCA method.
	}\label{table:result4}
\begin{tabular}{|c|c|c|c|}
  \hline
 & EX I&	EX II&	EX III\\
 \hline
ORL &	97\%&	84.64\%	&100\% \\
UMIST&	90.35\%	&85.18\%	&100 \% \\
JAFFE&	100\%	&100\%	&100\% \\
 \hline
\end{tabular}
\end{table}

\section{Conclusion}\label{sec:conclusion}
In this contribution, a power intelligent image processing system applied to recognition and classification of facial images is presented. Experimental results using the ORL, UMIST, JAFFE, and a new collected dataset for people in Hajj confirm the excellent performance of the new techniques presented in terms of recognition accuracy, speed, and storage requirements.


\end{document}